\documentclass[letterpaper]{article} 
\usepackage{aaai23}  
\usepackage{times}  
\usepackage{helvet}  
\usepackage{courier}  
\usepackage[hyphens]{url}  
\usepackage{graphicx} 
\usepackage{natbib}  
\usepackage{caption} 
\usepackage{tabularx}
\usepackage[table]{xcolor}  
\usepackage{centernot}
\usepackage{mathtools}
\usepackage{multirow}
\usepackage{array,colortbl}
\usepackage{tikz}
\usepackage{amsfonts}
\usepackage{float}
\usepackage{amsmath}
\usepackage{amssymb}
\usepackage{amsthm}
\usepackage{enumerate}
\usepackage{enumitem}
\usepackage{subcaption}
\usepackage[ruled,vlined,linesnumbered]{algorithm2e} 
\usepackage{newfloat}
\usepackage{listings}
\usepackage{booktabs}
\usepackage{xspace}
\SetNlSty{bfseries}{\color{black}}{}

\SetCommentSty{mycommfont}
\newcommand{\vv}{\ensuremath{\vec{v}}\xspace}
\newcommand{\uu}{\ensuremath{\vec{u}}\xspace}
\newcommand{\bb}{\ensuremath{\vec{b}}\xspace}

\def\CS{\mathrm{CS}}

\def\PCS{\mathrm{PCS}}
\def\CCS{\mathrm{CCS}}

\def\H{{\mathbf H}}

\def\w{{\vec{w}}}

\def\e{\varepsilon}

\DeclareMathOperator*{\argmin}{arg\,min}
\renewcommand{\phi}{\varphi}

\newcommand{\set}[1]{\left\{ #1 \right\}}
\newcommand{\seta}[1]{\{ #1 \}}

\newtheorem{assumption}{Assumption}
\newtheorem{theorem}{Theorem}[section]

\theoremstyle{definition}
\newtheorem{definition}[theorem]{Definition}

\def\R{{\mathbb R}}
\def\BB{{\mathrm{BellmanBackup}}}

\def\Q{{\mathbf Q}}
\def\V{{\mathbf V}}
\def\Vvec{{\vec{V}}}
\def\U{{\mathbf U}}

\def\C{{\vec{C}}}

\def\successors{\text{successors}}
\def\mossp{P=(S, s_0, G, A, P, \C)}

\newcommand{\heur}[2]{$\H_{\textit{#1}}^{\textit{#2}}$}
\newcommand{\hmax}{\ensuremath {h^{\mathsf{max}}}\xspace}

\newcommand{\comax}{\operatorname{comax}}

\newcommand{\rv}{\mathit{rv}}

\newcommand{\true}{\mathit{true}}
\newcommand{\closed}{\mathit{closed}}
\newcommand{\open}{\mathit{open}}
\newcommand{\solved}{\mathit{solved}}

\newcommand{\checkSolved}{\mathrm{checkSolved}}
\newcommand{\pop}{\mathit{pop}}

\SetKwProg{proc}{routine}{}{}
\SetKwProg{algo}{procedure}{}{}
\let\oldnl\nl
\newcommand{\nonl}{\renewcommand{\nl}{\let\nl\oldnl}}

\newcommand{\Omit}[1]{}

\newcommand{\oldnb}[1]{}

\newcommand{\PreserveBackslash}[1]{\let\temp=\\#1\let\\=\temp}
\newcolumntype{C}[1]{>{\PreserveBackslash\centering}p{#1}}
\newcolumntype{R}[1]{>{\PreserveBackslash\raggedleft}p{#1}}
\newcolumntype{L}[1]{>{\PreserveBackslash\raggedright}p{#1}}

\newcommand{\ProbVisitall}{Probabilistic VisitAll\xspace}

\newcommand{\VisitallWithFlat}{VisitAllTire\xspace}

\title{Heuristic Search for Multi-Objective Probabilistic Planning}
\author{
    Dillon Z. Chen,\textsuperscript{\rm 1}
    Felipe Trevizan,\textsuperscript{\rm 1}
    Sylvie Thi\'ebaux\textsuperscript{\rm 1,\rm 2}
}
\affiliations {
    \textsuperscript{\rm 1}School of Computing, The Australian National University\\
    \textsuperscript{\rm 2}LAAS-CNRS, ANITI, Universit\'e de Toulouse\\
    \{Dillon.Chen, Felipe.Trevizan, Sylvie.Thiebaux\}@anu.edu.au
}

\begin{document}

\maketitle

\begin{abstract}
    Heuristic search is a powerful approach that has successfully been applied to a broad class of planning problems, including classical planning, multi-objective planning, and probabilistic planning modelled as a stochastic shortest path (SSP) problem. Here, we extend the reach of heuristic search to a more expressive class of problems, namely multi-objective stochastic shortest paths (MOSSPs), which require computing a coverage set of non-dominated policies. We design new heuristic search algorithms MOLAO* and MOLRTDP, which extend well-known SSP algorithms to the multi-objective case. We further construct a spectrum of domain-independent heuristic functions differing in their ability to take into account the stochastic and multi-objective features of the problem to guide the search. Our experiments demonstrate the benefits of these algorithms and the relative merits of the heuristics.
\end{abstract}

\section{Introduction}
Stochastic shortest path problems (SSPs) are the de facto model for planning under uncertainty. Solving an SSP involves computing a policy which maps states to actions so as to minimise the expected (scalar) cost to reach the goal from a given initial state. Multi-objective stochastic shortest path problems (MOSSPs) are a useful generalisation of SSPs where multiple objectives (e.g. time, fuel, risk) need to be optimised without the user being able to \textit{a priori} weigh these objectives against each other \cite{roijers17modm}. In this more complex case, we now aim to compute a \emph{set} of non-dominated policies whose vector-valued costs represent all the possible trade-offs between the objectives.

There exist two approaches to solving MOMDPs \cite{roijers17modm}, a special case of MOSSPs: \emph{inner loop planning} which consists in extending SSP solvers by generalising single objective operators to the multi-objective (MO) case, and \emph{outer loop planning} which consists in solving a scalarised MOSSP multiple times with an SSP solver. We focus on the former. The canonical inner loop method, MO Value Iteration (MOVI) \cite{white82mdp} and its variants \cite{wiering07conmodp,barrett08learning,roijers17modm}, require enumerating the entire state space of the problem and are unable to scale to the huge state spaces typically found in automated planning. 

Therefore, this paper focuses on heuristic search methods which explore only a small fraction of the state space when guided with informative heuristics. Heuristic search has been successfully applied to a range of optimal planning settings, including single objective (SO) or constrained SSPs \cite{hansen01lao,bonet2003lrtdp,trevizan16idual}, and MO deterministic planning \cite{mandow2010namoa,Khouadjia2013,ulloa2020boastar}. Moreover, a technical report by Bryce \textit{et al.} \shortcite{bryce2007probabilistic} advocated the need for heuristic search and outlined an extension of LAO* \cite{hansen01lao} for finite horizon problems involving multiple objectives and partial observability. Convex Hull Monte-Carlo Tree-Search \cite{painter2020convex} extends the Trial-Based Heuristic Tree Search framework \cite{keller2013trial} to the MO setting but applies only to finite-horizon MOMDPs. Yet to the best of our knowledge there is no investigation of heuristic search for MOSSPs. 

This paper fills this gap in heuristic search for MOSSPs.
First, we characterise the necessary conditions for MOVI to converge in the general case of MOSSPs.
We then extend the well-known SSP heuristic search algorithms LAO* \cite{hansen01lao} and LRTDP \cite{bonet2003lrtdp} to the multi-objective case, leading to two new MOSSP algorithms, MOLAO* and MOLRTDP, which we describe along with sufficient conditions for their convergence.
We also consider the problem of guiding the search of these algorithms with domain-independent heuristics. A plethora of domain-independent heuristics exist for classical planning, but works on constructing heuristics for (single objective) probabilistic or multi-objective (deterministic) planning are much more recent \cite{trevizan17hpom,klossner2021pdb,geisser22moheuristics}.
Building on these recent works, we investigate a spectrum of heuristics for MOSSPs differing in their ability to account for the probabilistic and multi-objective features of the problem. 

Finally, we conduct an experimental comparison of these algorithms and of the guidance obtained via these heuristics. We observe the superiority of heuristic search over value iteration methods for MOSSPs, and of heuristics that are able to account for the tradeoffs between competing objectives. 

\section{Background}

A \emph{multi-objective stochastic shortest path problem (MOSSP)} is a tuple $(S, s_0, G, A, P, \C)$
where:
$S$ is a finite set of states, one of which is the initial state $s_0$,
$G \subseteq S$ is a set of goal states,
$A$ is a finite set of actions,
$P(s'\!\mid\!s,a)$ is the probability of reaching $s'$ after applying action $a$ in $s$, and
$\C(a) \in \R_{\ge 0}^n$ is the $n$-dimensional vector representing the cost of action $a$.
Two special cases of MOSSPs are stochastic shortest path problems (SSPs) and bi-objective SSPs which
are obtained when $n$ equals 1 and 2, respectively.

A solution for an SSP is a \emph{deterministic policy} $\pi$, i.e., a mapping from states to
actions.
A policy $\pi$ is \emph{proper} or \emph{closed w.r.t. $s_0$} if the probability of reaching $G$ when following $\pi$ from
$s_0$ is 1;
if this probability of reaching $G$ is less than 1, then $\pi$ is an improper policy.
We denote by $S^\pi \subseteq S$ the set of states visited when following a policy $\pi$ from
$s_0$.
The expected cost of reaching $G$ when using a proper policy $\pi$ from a state $s \in S^\pi$ is
given by the \emph{policy value function} defined as
\begin{align}\label{eq:v_pi_ssp}
V^\pi(s) = C(\pi(s)) + \sum_{s' \in S} P(s'|s,\pi(s)) V^\pi(s')
\end{align}
for $s \in S^\pi \setminus G$ and $V^\pi(g) = 0$ for $g \in G$.
An optimal policy for an SSP is any proper policy $\pi^*$ such that $V^{\pi^*}(s_0) \le
V^{\pi'}(s_0)$ for all proper policies $\pi'$.
Although $\pi^*$ might not be unique, the \emph{optimal value function} $V^*$ is unique and equals
$V^{\pi^*}$ for any $\pi^*$.

\emph{Stochastic policies} are a generalisation of deterministic policies which map states to
probability distributions of actions.
The definitions of proper, improper and policy value function are trivially generalised to
stochastic policies.
A key result for SSPs is that at least one of its optimal policies is
deterministic~\cite{bertsekas1991analysis};
thus it suffices to search for deterministic policies when solving SSPs.

\subsection*{Coverage sets and solutions for MOSSPs}
In the context of MOSSPs, given a policy $\pi$, we denote by $\Vvec^{\pi}:S \to \R_{\geq 0}^n$ the vector value function for $\pi$.
The function $\Vvec^{\pi}$ is computed by replacing $V$ and $C$ by $\Vvec$ and $\C$ in \eqref{eq:v_pi_ssp}, respectively.
In order to define the solution of an MOSSP, we need to first define how to compare two different vectors:
a cost vector \vv \emph{dominates} \uu, denoted as $\vv \preceq \uu$, if $\vv_i \leq \uu_i$ for $i=1,\ldots,n$.
A \emph{coverage set} for a set of vectors $\V$, denoted as $\CS(\V)$, is any set satisfying $\forall \vec{v} \in \CS(\V), \nexists \vec{u} \in \CS(\V)$ s.t. $\vec{u} \preceq \vec{v}$ and $\vec{u} \not= \vec{v}$.%
\footnote{We will denote sets of vectors or functions which map to sets of vectors with bold face, e.g. $\V$, and single vectors or functions which map to single vectors with vector notation, e.g. $\vec{V}$.}
An example of a coverage set is the \emph{Pareto coverage set} ($\PCS$) which is the largest possible coverage set. For the remainder of the paper we focus on the \emph{convex coverage set} ($\CCS$)~\cite{barrett08learning,roijers17modm} which is defined as the convex hull of the $\PCS$. Details for computing the $\CCS$ of a set $\V$ with a linear program (LP) can be found in~\cite[Sec. 4.1.3]{roijers17modm}. We say that a set of vectors $\U$ \emph{dominates} another set of vectors $\V$, denoted by $\U \preceq \V$, if for all $\vec{v} \in \V$ there exists $\vec{u} \in \U$ such that $\vec{u} \preceq \vec{v}$.

Given an MOSSP define the \emph{optimal value function} $\V^*$ by $\V^*(s) = \CCS(\seta{\Vvec^{\pi}(s) \mid \text{$\pi$ is a proper policy}})$. 
Then we define a \emph{solution} to the MOSSP to be any set of proper policies $\Pi$ such that the function $\phi:\Pi \to \V^*(s_0)$ with $\phi(\pi) = \vec{V}^{\pi}(s_0)$ is a bijection.
By choosing $\CCS$ as our coverage set operator, we may focus our attention to only non-dominated deterministic policies, where non-dominated stochastic policies are implicitly given by the points on the surface of the polyhedron drawn out by the $\CCS$.
In this way, we avoid having to explicitly compute infinitely many non-dominated stochastic policies. 

\begin{figure}[t]
    \centering
    \begin{tikzpicture}
        \node[shape=circle,draw=black,fill=blue!5] (A) at (0,0) {$s_0$};
        \node[shape=circle,draw=black,minimum size=0.75cm] (B) at ([shift={(-2.5,0)}]A) {$g_1$};
        \node[shape=circle,draw=black,minimum size=0.75cm] (C) at ([shift={(2.5,0)}]A) {$g_2$};
        \node[shape=circle,draw=black,minimum size=0.6cm] (B2) at (B) {};
        \node[shape=circle,draw=black,minimum size=0.6cm] (C2) at (C) {};
        \path [->] (A) edge node[below] {$a_1$} (B);
        \path [->] (A) edge node[below] {$a_2$} (C);
        \coordinate (L) at ([shift={(-1,0)}]A) {}; 
        \coordinate (R) at ([shift={( 1,0)}]A) {}; 
        \path [->] (L) edge[bend left=60] (A);
        \path [->] (R) edge[bend right=60] (A);
        \node[] (Bx) at ([shift={( 0.75,0.25)}]B) {$0.5$};
        \node[] (Cx) at ([shift={(-0.75,0.25)}]C) {$0.5$};
        \node[] (Al) at ([shift={(-0.55,0.60)}]A) {$0.5$};
        \node[] (Ar) at ([shift={( 0.55,0.60)}]A) {$0.5$};
    \end{tikzpicture}
    \caption{A MOSSP with action costs given by $\vec{C}(a_1) = [1,0]$ and $\vec{C}(a_2) = [0, 1]$.}
    \label{fig:infinite}
\end{figure}
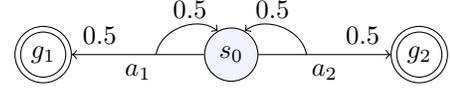

To illustrate this statement, consider the MOSSP in Fig.~\ref{fig:infinite} with actions $a_1$ and $a_2$ where $P(s_0|s_0,a_1)=P(g_1 | s_0, a_1)=P(s_0|s_0,a_2)=P(g_2 |s_0, a_2)=0.5$ and $\vec{C}(a_1)=[1,0]$ and $\vec{C}(a_2)=[0,1]$.
One solution consists of only two deterministic policies $\pi_1(s_0)=a_1$ and $\pi_2(s_0)=a_2$ with corresponding expected costs ${[2,0]}$ and ${[0,2]}$.
Notice that there are uncountably many stochastic policies obtained by the convex combinations of $\pi_1$ and $\pi_2$, i.e., $\pi_t(a_1 | s_0) = 1-t, \pi_t(a_2 | s_0) = t$ for $t \in [0,1]$.
The expected cost of each $\pi_t$ is $[2-2t,2t]$ and these stochastic policies do not dominate each other.
Therefore, if the $\PCS$ is used instead of the $\CCS$ in the definition of optimal value function, then $\V^*(s_0)$ would be $\set{[2-2t,2t] \mid t\in[0,1]}$ and the corresponding solution would be $\{\pi_t \mid t \in [0,1]\}$.
The $\CCS$ allows us to compactly represent the potentially infinite $\PCS$ by storing only the deterministic policies: in this example, the actual solution is $\{\pi_1,\pi_2\}$ and the optimal value function is $\V^*(s_0)=\set{[2,0],[0,2]}$.

\subsection*{Value Iteration for SSPs}

We finish this section by reviewing how to compute the optimal value function ${V}^*$ for SSPs and
extend these results to MOSSPs in the next section.
The optimal (scalar) value function ${V}^*$ for an SSP can be computed using the Value Iteration (VI)
algorithm~\cite{bertsekas96neuro}: given an initial value function ${V}^0$ it computes the sequence
${V}^1, \ldots, {V}^k$ where ${V}^{t+1}$ is obtained by applying a Bellman backup, that is ${V}^{t+1}(g)=0$ if $g \in G$ and, for $s \in S \setminus G$,
\begin{align}
{V}^{t+1}(s) &= \min_{a \in A} {Q}^{t+1}(s, a) \label{eq:bb1} \\
{Q}^{t+1}(s, a) &= {C}(a) + \sum_{s' \in S} P(s'|s, a){V}^{t}(s'). \label{eq:ssp_qvalue}
\end{align}
VI guarantees that ${V}^k$ converges to ${V}^*$ as $k \to \infty$ under the following conditions:
(i) for all $s \in S$, there exists a proper policy w.r.t. $s$; and
(ii) all improper policies have infinite cost for at least one state.
The former condition is known as the \emph{reachability assumption} and it is equivalent to requiring
that no dead ends exist, while the latter condition can be seen as preventing cycles with 0 cost.
Notice that \emph{finite-horizon Markov Decision Processes} (MDPs) and \emph{infinite-horizon discounted MDPs}
are special cases of SSPs in which all policies are guaranteed to be proper~\cite{bertsekas96neuro}.

\section{Value Iteration for MOSSPs}\label{sec:movi}

Value Iteration has been extended to the MO setting in special cases of SSPs, e.g.
infinite-horizon discounted MDPs~\cite{white82mdp}. In this section, we present the MO version of Value Iteration for the
general MOSSP case.
While the changes in the algorithm are minimal, the key contribution of this section is the
generalisation of the assumptions needed for convergence.
We start by generalising the Bellman backup operation from the SO, i.e. \eqref{eq:bb1} and
\eqref{eq:ssp_qvalue}, to the MO case. For $s \in S \setminus G$ we have
\begin{align}
    \V^{t+1}(s) &= \CS\Big(\bigcup_{a \in A} \Q^{t+1}(s, a)\Big) \label{eq:mobb1}  \\
    \Q^{t+1}(s, a) &= \seta{\C(a)} \oplus \Big( \bigoplus_{s' \in S} P(s'|s, a) \V^{t}(s') \Big)
    \label{eq:mo_q_val},
\end{align}
and $\V^{t+1}(g) = \{\vec{0}\}$ for $g \in G$ where $\oplus$ denotes the sum of two sets of vectors $\V$
and $\U$ defined as $\{ \uu + \vv \mid \uu \in \U, \vec{v} \in \V\}$, and $\bigoplus$ is the
generalised version of $\oplus$ to several sets. 

\begin{algorithm}[t]
  \caption{\textsc{MOVI}}\label{alg:movi}
  \KwData{MOSSP problem $\mossp$, initial values $\V(s)$ for each state $s$ (default to $\V(s) = \{\vec{0}\}$), and consistency threshold $\e$.}
  \While{$\max_{s \in S} \textit{res}(s) < \e$}{ \label{line:movi:conv}
      \For{$s \in S$}{  \label{line:movi:start_bb}
          \lIf{$s \in G$}{
              $\V_{\textit{new}}(s) \leftarrow \{\vec{0}\}$
          }\lElse{
              $\V_{\textit{new}}(s) \leftarrow \BB(s)$
          }  \label{line:movi:end_bb}
          $\textit{res}(s) \leftarrow D(\V, \V_{\textit{new}})$ \label{line:movi:res}\\
      }
      $\V \leftarrow \V_{\textit{new}}$\\
  }
  \Return{$\V$}\\
\end{algorithm}

Alg. \ref{alg:movi} {illustrates the MO version of Value Iteration (MOVI) which is very similar to the single-objective VI algorithm with the notable difference that
the convergence criterion is generalised to handle sets of vectors.}
The Hausdorff distance between two sets of vectors $\U$ and $\V$ is given by 
\begin{align}
    \hspace*{-0.15cm}D(\U, \V) = \max\!\big\{\!\max_{\vec{u} \in \U} \min_{\vec{v} \in \V} d(\vec{u}, \vec{v}), \max_{\vec{u} \in \V} \min_{\vec{v} \in \U} d(\vec{u}, \vec{v})\big\}
\end{align}
for some choice of metric $d$ such as the Euclidean metric. We use this distance to define residuals in line \ref{line:movi:res}.
As with VI, MOVI converges to the optimal value function at the limit~\cite{white82mdp,barrett08learning} under certain strong assumptions presented in the next section. We can extract policies with the choice of a scalarising weight $\w$ from the value function~\cite{barrett08learning}.

\begin{algorithm}[t]
  \caption{\textsc{MOVI} under Assumption 1}\label{alg:moviass1}
  \KwData{MOSSP problem $\mossp$, initial values $\V(s)$ for each state $s$ (default to $\V(s) = \{\vec{0}\}$), consistency threshold $\e$ and upper bound \bb.}
  \While{$\max_{s \in S} \textit{res}(s) < \e$}{
      \For{$s \in S$}{
          \lIf{$s \in G$}{
              $\V_{\textit{new}}(s) \leftarrow \{\vec{0}\}$
          }\lElse{
              $\V_{\textit{new}}(s) \leftarrow \BB_B(s)$ \label{line:moviass1:modified}
          }
          $\textit{res}(s) \leftarrow D(\V, \V_{\textit{new}})$\\
      }
      $\V \leftarrow \V_{\textit{new}}$\\
  }
  \lFor{$s \in S$}{
      $\V(s) = \V(s)\setminus\seta{\bb}$ \label{line:moviass1:prune}
  }
  \Return{$\V$}\\
\end{algorithm}

\subsection*{{Assumptions for the convergence of MOVI}}

Similarly to the SO version of VI, MOVI requires that the reachability assumption holds; however, the
assumption that improper policies have infinite cost needs to be generalised to the MO case.
One option is to require that all improper policies cost $\vec{\infty}$ which we call the \textbf{strong improper policy assumption} and, if it holds with the reachability assumption, then MOVI (Alg.~\ref{alg:movi}) is sound and complete for MOSSPs.
The strong assumption is very restrictive because it implies that any cycle in an MOSSP must have a cost greater than zero in \textit{all dimensions};
however, it is common for MOSSPs to have zero-cost cycles in one or more dimensions.
For instance, in navigation domains where some actions such as wait, load and unload do not consume fuel and can be used to generate zero-fuel-cost loops.

\begin{figure}[b]
    \centering
    \begin{tikzpicture}
        \node[shape=circle,draw=black,fill=blue!5] (A) at (0,0) {$s_0$};
        \node[shape=circle,draw=black] (B) at ([shift={(-2.5,0)}]A) {$s_1$};
        \node[shape=circle,draw=black,minimum size=0.75cm] (C) at ([shift={(2.5,0)}]A) {$g$};
        \node[shape=circle,draw=black,minimum size=0.6cm] (B2) at (C) {};
        \path [<-] (A) edge[bend left] [above] node {$a_2$} (B);
        \path [->] (A) edge[bend right] [below] node {$a_1$} (B);
        \path [->] (A) edge [above] node {$a_g$} (C);
    \end{tikzpicture}
    \caption{An MOSSP with action costs given by $\vec{C}(a_1) = [1,0], {\vec{C}(a_2) = [1, 0]}, \vec{C}(a_g) = [0,1]$.}
    \label{fig:vi-fails}
\end{figure}
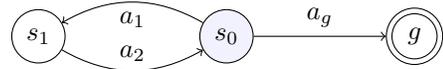

Another possible generalisation for the cost of improper policies is that they cost infinity in \emph{at least one dimension}.
This requirement is not enough as illustrated by the (deterministic) MOSSP in Fig.~\ref{fig:vi-fails}.
The only deterministic proper policy is $\pi(s_0) = a_g$ with cost $[0,1]$, and the other deterministic policy is $\pi'(s_0)=a_1, \pi'(s_1)=a_2$ which is improper and costs $[\infty,0]$.
Since neither $[0,1]$ or $[\infty,0]$ dominate each other, two issues arise:
MOVI tries to converge to infinity and thus never terminates; and
even if it did, the cost of an improper policy would be wrongly added to the CCS.

These issues stem from VI and its generalisations not explicitly pruning improper policies.
Instead they rely on the assumption that improper policies have infinite cost to implicitly prune them.
This approach is enough in the SO case because any proper policy dominates all improper policies since domination is reduced to the `$\le$' operator;
however, as shown in this example, the implicit pruning approach for improper policies is not correct for the MO case.

We address these issues by providing a version of MOVI that explicitly prunes improper policies and
to do so we need a new assumption:

\begin{assumption}[Weak Improper Policy Assumption]\label{assump:weak}
There exists a vector $\vec{b} \in \R_{\geq 0}^n$ such that for every improper policy $\pi$ and $s \in S$, $\Vvec^{\pi}(s) \not\preceq \vec{b}$, and for every proper policy $\pi$ and $s \in S$, $\vec{V}^\pi(s) \preceq \bb$ and $\vec{V}^\pi(s) \neq \bb$.
\end{assumption}

The weak assumption allows improper policies to have finite cost in some \emph{but not all} dimensions at the expense of knowing an upper bound \bb on the cost of all proper policies.
This upper bound \bb lets us infer that a policy $\pi$ is improper whenever $\vec{V}^\pi(s) \not\preceq \bb$, i.e., that $\vec{V}^\pi(s)$ is greater than \bb in at least one dimension.

Alg.~\ref{alg:moviass1} outlines the new MOVI where the differences can be found in lines \ref{line:moviass1:modified} and \ref{line:moviass1:prune}.
In line \ref{line:moviass1:modified}, we use a modified Bellman backup which detects vectors associated to improper policies and assigns them the upper bound \bb given by the weak improper policy assumption.
The modified backup is given by
\begin{align}
    \V^{t+1}(s) &= \CS_B\biggl(\bigcup_{a \in A} \Q_B^{t+1}(s, a)\biggr) \label{eqn:newbb}
    \\
    \Q_B^{t+1}(s, a) &= \seta{\C(a)} \oplus_B \Big( \bigoplus_{s' \in S}\!\!\!\!\!\!\!\!\!\!{\phantom{{a_3}_3}}_B P(s'|s, a) \V^{t}(s') \Big)
\end{align}
where $\oplus_B$ is a modified set addition operator defined on pairwise vectors $\vec{u}$ and $\vec{v}$ by
\begin{align} \label{eqn:oplusB}
    \vec{u} \oplus_B \vec{v} =
    \begin{cases}
        \bb &\text{if $\vec{u} \oplus \vec{v} \not\preceq \bb$} \\
        \vec{u} \oplus \vec{v} &\text{otherwise},
    \end{cases}
\end{align}
and similarly for the generalised version ${{\bigoplus}}_B$.
Also define $\CS_B$ to be the modified coverage set operator which does not prune away $\bb$ if it is in the input set.
The modified backup \eqref{eqn:newbb} enforces a cap on value functions (i.e., $\V^{t}(s) \preceq \{\bb\}$) through the operator~$\oplus_B$~\eqref{eqn:oplusB}.
This guarantees that a finite fixed point exists for all $\vec{V}^\pi(s)$ and, as a result, that Alg.~\ref{alg:moviass1} terminates.
Once the algorithm converges, we need to explicitly prune the expected cost of improper policies which is done in line \ref{line:moviass1:prune}.
By Assumption~\ref{assump:weak}, we have that no proper policy costs \bb thus we can safely remove \bb from the solution.
Note that the overhead in applying this modified backup and post-processing pruning is negligible.
Theorem~\ref{thm:movi} shows that MOVI converges to the correct solution.

\begin{theorem}\label{thm:movi}
Given an MOSSP in which the reachability and weak improper policy assumptions hold for an upper
bound \bb, and given a set of vectors $\V^0$ such that
$\V^0 \preceq \V^*$,
the sequence $\V^1, \ldots, \V^k$ computed by MOVI converges to the MOSSP solution $\V^*$ as $k
\to \infty$.
\end{theorem}
\begin{proof}[Proof sketch]

By the assumption that $\V^0 \preceq \V^*$, we have that MOVI will not prune away vectors associated with proper policies which contribute to a solution.
If $\V^0 \not\preceq \V^*$, e.g., $\V^0(s) = \{\bb\}$ for all $s$, then Alg.~\ref{alg:moviass1} is not guaranteed to find the optimal value function since it will incorrectly prune proper policies.
Otherwise, we have that $\V^1, \ldots, \V^k$ converges to $\V^{\dagger} = \V^* \cup \seta{\bb}$ if the original MOVI in Alg. \ref{alg:movi} does not converge, and $\V^{\dagger} = \V^*$ otherwise.
For example, $\V^*=\set{[0,1]}$ in the example seen in Fig.~\ref{fig:vi-fails} but $\V^{\dagger} = \set{[2, 2], [0, 1]}$ if we choose $\bb=[2,2]$.

This convergence result follows by noticing that, by definition of the modified backup in \eqref{eqn:newbb}, every vector in $\V^t(s)$ for all $t$ dominates $\bb$.
We may then apply the proof for the convergence of MOVI with convex coverage sets by Barrett and Narayanan \shortcite{barrett08learning} which reduces to the convergence of scalarised SSPs using VI in the limit, of which there are finitely many since the number of deterministic policies is finite.
Here, we have for every $\vec{w}$ that $\vec{w}\cdot\bb$ is an upper bound for the problem scalarised by $\vec{w}$. Finally, we have that line 7 removes $\bb$ from $\V^{\dagger}$ such that we correctly return $\V^*$.
\end{proof}

Note that the original proof for MOMDPs by Barrett and Narayanan \shortcite{barrett08learning} does not directly work for MOSSPs as some of the scalarised SSPs have a solution of infinite expected cost such that VI never converges. The upper bound \bb we introduce solves this issue as achieving this bound is the same as detecting improper policies. The proof remains correct in the original setting applied to discounted MOMDPs in which there are no improper policies.

\subsection*{{Relaxing the reachability assumption}}
The reachability assumption can also be relaxed by transforming an MOSSP with dead ends into a new
MOSSP without dead ends.
Formally, given an MOSSP $(S, s_0, G, A, P, \C)$ with dead ends, let
$A' = A \cup \{\text{give-up}\}$;
$P(s_g|\text{give-up},s) = 1$ for all $s \in S$ and any $s_g \in G$;
$\C'(a) = [\C(a): 0]$ for all $a \in A$; and
$\C'(\text{give-up}) = [\vec{0}: 1]$.

Then the MOSSP $(S, s_0, G, A', P, \C')$ does not have dead ends (i.e., the reachability assumption
holds) since the give-up action is applicable in all states $s \in S$.
This transformation is similar to the \emph{finite-penalty transformation} for
SSPs~\cite{trevizan17:mcmp}; however it does not require a large penalty constant.
Instead, the MOSSP transformation uses an extra cost dimension to encode the cost of giving up and
the value of the $n\!+\!1$-th dimension of $\C'(\text{give-up})$ is irrelevant as long as it is
greater than 0.
Since the give-up action can only be applied once, defining the cost of give-up as 1 let us interpret
the $n\!+\!1$-th dimension of any $\vec{V}^\pi(s)$ as the probability of giving up when following $\pi$
from $s$.

\section{Heuristic search}
Value iteration gives us one method for solving MOSSPs but is limited by the fact that it requires enumerating over the whole state space. This is impractical for planning problems,  as their state space grows exponentially with the size of the problem encoding.
This motivates the need for heuristic search algorithms in the vein of
LAO$^*$ and LRTDP for SSPs \cite{hansen01lao,bonet2003lrtdp}, which perform backups on a promising subset of states at each iteration.
To guide the choice of the states to expand and explore at each iteration, they use heuristic estimates of state values initialised when the state is first encountered.
In this section, we extend the concept
of heuristics to the more general MOSSP setting, {discuss a range of ways these heuristics can be constructed},  and provide multi-objective versions of LAO$^*$ and LRTDP.

\subsection{Heuristics}
For the remainder of the paper, we will call a set of vectors a value. Thus, a heuristic value for a state is a set of vectors $\H(s) \subset \R_{\geq 0}^n$. We implicitly assume that any heuristic value $\H(s)$ has already been pruned with some coverage set operator. The definition of an admissible heuristic for MOSSPs should be at most as strong as the definition for deterministic MO search \cite{mandow2010namoa}.
\begin{definition}
    A heuristic $\H$ for an MOSSP is \emph{admissible} if $\forall s \in S\setminus G, \H(s) \preceq \V^*(s)$ where $\V^*$ is the optimal value function, and $\forall g \in G, \H(g) = \seta{\vec{0}}$.
\end{definition}
For example, if for some MOSSP we have $\V^*(s)=\set{[0,2]}$ for a state $s$, then $\H(s)=\set{[0,1], [3,0]}$ and $\H(s')=\seta{\vec{0}}$ for all other $s'$ is an admissible heuristic.
As in the single-objective case, admissible heuristics ensure that the algorithms below converge to near optimal value functions for $\e>0$ with finitely many iterations and to optimal value functions with possibly infinitely many iterations when $\e=0$.

\begin{definition}
    A heuristic $\H$ for an MOSSP is \emph{consistent} if we have for all $s \in S, a \in A$ that $\H(s) \preceq  \seta{\C(a)} \oplus (\bigoplus_{s' \in S} P(s'|s, a) \H(s')).$
\end{definition}
The definition of consistent heuristic is derived and generalised from the definition of consistent heuristics for deterministic search: $h(s) \leq c(s,a,s') + h(s')$.
The main idea is that assuming non-negative costs, state costs increase monotonically which results in no re-expansions of nodes in A$^*$ search.
Similarly, we desire monotonically increasing value functions for faster convergence.

We now turn to the question of constructing domain-independent heuristics satisfying these properties from the encoding of a planning domain. This question has only recently been addressed in the \textit{deterministic} multi-objective planning setting \cite{geisser22moheuristics}: with the exception of this latter work, MO heuristic search algorithms have been evaluated using ``ideal-point'' heuristics, which apply a single-objective heuristic $h_i$ to each objective in isolation, resulting in a single vector $\H_{\textit{ideal}}(s) = \{[h_1(s), \ldots h_n(s)]\}$.  Moreover, informative domain-independent heuristics for single objective SSPs are also relatively new \cite{trevizan17hpom,klossner2021pdb}: a common practice was to resort to classical planning heuristics obtained after determinisation of the SSP \cite{jimenez06determinisation}.

We consider a spectrum of options when constructing domain-independent heuristics for MOSSPs, which we instantiate using the most promising families of heuristics identified in \cite{geisser22moheuristics} for the deterministic case: critical paths and abstraction heuristics. All heuristics are consistent and admissible unless otherwise mentioned.

\begin{itemize}
    \item The baseline option is to ignore both the MO and stochastic aspects of the problem, and resort to an ideal-point heuristic constructed from the determinised problem. In our experiments below,
    we apply the classical $\hmax$ heuristic \cite{bonet2001heuristics} to each objective and call this \heur{ideal}{max}.
    \item A more elaborate option is to consider only the stochastic aspects of the problem, resulting in an ideal-point SSP heuristic. In our experiments, we apply the recent SSP canonical PDB abstraction heuristic by Kl\"o{\ss}ner and Hoffmann \shortcite{klossner2021pdb} to each objective which we call \heur{ideal}{pdb2} and \heur{ideal}{pdb3} for patterns of size 2 and 3, respectively.
    \item Alternatively, one might consider only the multi-objective aspects, by applying some of the MO deterministic heuristics \cite{geisser22moheuristics} to the determinised SSP. The heuristics we consider in our experiments are the MO extension of \hmax and canonical PDBs: \heur{mo}{comax}, \heur{mo}{pdb2}, and \heur{mo}{pdb3}. These extend classical planning heuristics by using MO deterministic search to solve subproblems and combining solutions by taking the maximum of two sets of vectors with the admissibility preserving operator $\comax$ \cite{geisser22moheuristics}. 
    \item The best option is to combine the power of SSP and MO heuristics. We do so with a novel heuristic \heur{mossp}{pdb} which extends the SSP PDB abstraction heuristic \cite{klossner2021pdb} to the MO case by using an MO extension of topological VI (TVI) \cite{dai2014topological} to solve each projection and the $\comax$ operator to combine the results. Our experiments use \heur{mossp}{pdb2} and \heur{mossp}{pdb3}.
\end{itemize}

\subsection{(i)MOLAO$^*$}

Readers familiar with the LAO$^*$ heuristic search algorithm and the multi-objective backup can convince themselves that an extension of LAO$^*$ to the multi-objective case can be obtained by replacing the backup operator with the multi-objective version.
This idea was first outlined by Bryce, Cushing, and Kambhampati \shortcite{bryce2007probabilistic} for \emph{finite-horizon MDPs} which is a special case of SSPs without improper policies.
In the same vein as \cite{hansen01lao}, we provide MOLAO$^*$ alongside an `improved' version, iMOLAO$^*$, in Alg. \ref{alg:molao} and \ref{alg:imolao} respectively.

MOLAO$^*$ begins in line 1 by lazily assigning an initial value function $\V$ to each state with the heuristic function, as opposed to explicitly initialising all initial values at once. $\Pi$ is a dictionary representing our partial solution which maps states to sets of optimal actions corresponding to their current value function. 
The main while loop of the algorithm terminates once there are no nongoal states on the frontier as described in line \ref{line:molao_conv}, at which point we have achieved a set of closed policies.
\oldnb{``Closed'' is defined already, so I rewrote the sentence. Also shouldn't we say something stronger than just ``a set of closed policies''? Maybe ``a set of closed policies that is the optimal solution to the MOSSP''?}
\oldnb{DC: I think what you have written is fine. We mention later the algorithm is indeed correct and complete.}

The loop begins with lines 3-5 by removing a nongoal state $s$ from the frontier representing a state on the boundary of the partial solution graph, and adding it to the interior set $I$. The nodes of the partial solution graph are partial solution states, and the edges are the probabilistic transitions under all partial solution actions. Next in line 6 we add the set of 
yet unexplored successors of $s$ to the frontier: any state $s' \in S \setminus I$ such that $\exists a \in A, P(s'|s,a) > 0$.
Then we extract the set $Z$ of all ancestor states of $s$ in the partial solution $\Pi$ using graph search in line 7.
We run MOVI to $\e$-consistency on the MOSSP problem restricted to the set of states $Z$ and update the value functions for the corresponding states in line 8. The partial solution is updated in line 9 by extracting the actions corresponding to the value function with
\begin{align}
    \hspace*{-0.1cm}\mathrm{getActions}(s, \V) = \set{a \in A \mid \Q(s, a) \cap \V(s) \not= \emptyset}. \label{eqn:getactions}
\end{align}
This function can be seen as a generalisation of $\argmin$ to the MO case where here we select the actions whose $\Q$ value at $s$ contribute to the current value $\V(s)$.
Next, we extract the set of states $N$ corresponding to all states reachable from $s_0$ by the partial solution $\Pi$ in line 10.

\begin{algorithm}[t]
  \caption{\textsc{MOLAO$^*$}}\label{alg:molao}
  \KwData{MOSSP problem $\mossp$, heuristic $\H$, and consistency threshold $\e$}
  $\V \leftarrow \H; \;
  \Pi \leftarrow \emptyset; \;
  F \leftarrow \set{s_0}; \;
  I \leftarrow \emptyset; \;
  N \leftarrow \set{s_0}$ \\
  \While{$(F \cap N) \setminus G \not= \emptyset$}{ \label{line:molao_conv}
      $s \leftarrow \text{any element from $(F \cap N) \setminus G$}$\\
      $F \leftarrow F \setminus \set{s}$\\
      $I \leftarrow I \cup \set{s}$\\
      $F \leftarrow F \cup (\successors(s) \setminus I)$\\
      $Z \leftarrow \mathrm{ancestorStates}(s, \Pi)$\\
      $\V|_Z \leftarrow \text{MOVI}(P|_Z, \V|_Z, \e)$\\
      \lFor{$s \in Z$}{
          $\Pi(s) \leftarrow \mathrm{getActions}(s, \V)$
      }
      $N \leftarrow \text{solutionGraph}(s_I, \Pi)$\\
  }
  \Return{$\V$}\\
\end{algorithm}

The completeness and correctness of MOLAO$^*$ follows from the same reasoning as LAO$^*$ extended to the MO case. Specifically, we have that given an admissible heuristic the algorithm achieves $\e$-consistency upon termination.

\begin{algorithm}[t]
  \caption{\textsc{iMOLAO$^*$}}\label{alg:imolao}
  \KwData{MOSSP problem $\mossp$, heuristic $\H$, and consistency threshold $\e$}
  $\V \leftarrow \H; \;
  \Pi \leftarrow \emptyset; \;
  F \leftarrow \set{s_0}; \;
  I \leftarrow \emptyset; \;
  N \leftarrow \set{s_0}$ \\
  \While{$((F \cap N) \setminus G \not= \emptyset) \wedge (\max_{s \in N} \textit{res}(s) < \e)$}{
      $F = \emptyset$\\
      $N \leftarrow \text{postorderTraversalDFS}(s_I, \Pi)$\\
      \For{$s \in N$ in the computed order}{
          $\V(s) \leftarrow \BB(s)$\\
          $\Pi(s) = \mathrm{getActions}(s, \V)$\\
          \lIf{$s \notin I$}{
              $F = F \cup \set{s}$
          }
          $I = I \cup \set{s}$ \\
      }
  }
  \Return{$\V$}\\
\end{algorithm}

One potential issue with MOLAO$^*$ is that we may waste a lot of backups while running MOVI to convergence several times on partial solution graphs which do not end up contributing to the final solution.
The original authors of LAO$^*$ proposed the iLAO$^*$ algorithm to deal with this.
We provide the MO extension, namely iMOLAO$^*$.
The main idea with iMOLAO$^*$ is that we only run one set of backups every time we (re-)expand a state instead of running VI to convergence in the loop in lines 5 to 9. Backups are also performed asynchronously using DFS postorder traversal of the states in the partial solution graph computed in line 4, allowing for faster convergence times.

\begin{algorithm}[!ht]
  \caption{\textsc{MOLRTDP}}\label{alg:molrtdp}
  \KwData{MOSSP problem $\mossp$, heuristic $\H$, and consistency threshold $\e$}
  \nonl\algo{$\textsc{MOLRTDP}(P, \e, \H)$}{
      \setcounter{AlgoLine}{0}
      $\V \leftarrow \H$ \\
      \While{$\neg s_0.\solved$}{
          $\textit{visited} \leftarrow \emptyset$\\
          $s \leftarrow s_0$\\
          \While{$\neg s.\solved$}{
              $\textit{visited.push}(s)$\\
              \lIf{$s \in G$}{
                  \textbf{break}
              }
              $\V(s) \leftarrow \mathrm{BellmanBackup}(s)$\\
              $a \leftarrow \mathrm{sampleUnsolvedGreedyAction}(s)$\\
              $s \leftarrow \mathrm{sampleUnsolvedNextState}(s, a)$\\
          }

          \While{$\neg\textit{visited.empty()}$}{
              $s \leftarrow \textit{visited.pop()}$\\
              \lIf{$\neg\checkSolved(s)$}{
                  \textbf{break}
              }
          }
      }
      \Return{$\V$}\\
  }
  \nonl\proc{$\checkSolved(s)$}{
      \setcounter{AlgoLine}{0}
      $\rv \leftarrow \true;\; 
      \open \leftarrow \emptyset;\; 
      \closed \leftarrow \emptyset$\\
      \lIf{$\neg s.\solved$}{
          $open.\mathrm{push}(s)$
      }

      \While{$\neg\textit{open.empty}()$}{
          $s \leftarrow \textit{open.pop}()$\\
          \If{$\textit{res}(s) > \e$}{
              $\mathit{rv} \leftarrow \mathit{false}$\\
              \textbf{continue}\\
          }
          \For{$a \in \mathrm{getActions}(s, \V)$}{
              \For{$s' \in \mathrm{successors}(s, a)$}{
                  \lIf{$\neg s'.\solved \wedge s' \notin \open \cup \closed$}{
                      $\textit{open.push}(s')$
                  }
              }
          }
      }

      \lIf{$\rv$}{
          \textbf{for} $s \in \closed$ \textbf{do} $s.\solved = \true$
      }
      \Else{
          \While{$\closed\not=\emptyset$}{
              $s \leftarrow \closed.\pop()$\\
              $\V(s) \leftarrow \BB(s)$
          }
      }
      \Return{$\rv$}\\
  }
\end{algorithm}

To summarise, the two main changes required to extend (i)LAO$^*$ to the MO case are (1) replacing the backup operator with the MO version, and (2) storing possibly more than one greedy action at each state corresponding to incomparable vector $\Q$-values, resulting in a larger set of successors associated to each state. These ideas can be applied to other asynchronous VI methods such as Prioritised VI \cite{wingate2005prioritization} and Topological VI \cite{dai2014topological}.

\subsection{MOLRTDP}

LRTDP \cite{bonet2003lrtdp} is another heuristic search algorithm for solving SSPs. The idea of LRTDP is to perform random trials using greedy actions based on the current value function or heuristic for performing backups, and labelling states as solved when the consistency threshold has been reached in order to gradually narrow down the search space and achieve
a convergence criterion. A multi-objective extension of LRTDP is not as straightforward as extending the backup operator given that each state has possibly more than one greedy action to account for. The two main changes from the original LRTDP are (1) the sampling of a random greedy action before sampling successor states in the main loop, and (2) defining the descendants of a state by considering successors of all greedy actions (as opposed to just one in LRDTP). Alg. \ref{alg:molrtdp} outlines the MO extension of LRTDP.

MOLRTDP consists of repeatedly trialing paths through the state space and performing backups until the initial state is marked as solved. Trials are run by randomly sampling a greedy action $a$ from $\mathrm{getActions}(s, \V)$ at each state $s$ followed by a next state sampled from the probability distribution of $a$ until a goal state is reached as in the first inner loop from lines 5 to 10. The second inner loop from lines 11 to 13 calls the $\checkSolved$ routine in the reverse trial order to label states as solved by checking whether the residual of all its descendant states under greedy actions are less than the convergence criterion. 

The $\checkSolved$ routine begins with inserting $s$ into the open set if it has not been solved, and returns true otherwise due to line 2. The main loop in lines 3 to 10 collects the descendent states under all greedy actions (lines 8-10) and checks whether the residual of all the descendent states are small (lines 5-7). If true, all descendents are labelled as solved as in line 11. Otherwise, backups are performed in the reverse order of explored states as in lines 12 to 15.

The completeness and correctness of MOLRTDP follows similarly from its single objective ancestor. We note specifically that the labelling feature works similarly in the sense that whenever a state is marked as solved, it is known for certain that all its descendents' values are $\e$-consistent and remain unchanged when backups are performed elsewhere.

\section{Experiments}

In this section we empirically evaluate the different algorithms and heuristics for MOSSPs in
several domains.
Since no benchmark domains for MOSSPs exist, we adapt domains from a variety of sources to
capture challenging features of both SSPs and MO deterministic planning.
Our benchmark set is introduced in the next section and it is followed by our experiments setup and
analysis of the results.

\subsection{Benchmarks}\oldnb{FT: if we have space, it would be nice to add the param of each domain and the range used}

\subsubsection{$k$-d Exploding Blocksworld} Exploding Blocksworld (ExBw) was first introduced by Younes et al. \shortcite{younes2005competition} and later slightly modified for the IPPC'08~\cite{bryce08ippc}. In ExBw, a robot has to pick up and stack blocks on top of each other to get a target tower configuration. Blocks have a chance of detonating and destroying blocks underneath or the table. We consider a version with no dead ends using an action to repair the table~\cite{trevizan16idual}. We extend ExBw to contain multi-objective costs. ExBw-2d has two objectives: the number of actions required to stack the blocks and number of times we need to repair the table. ExBw-3d has an additional action to repair blocks with an objective to minimise the number of block repairs.

\subsubsection{MO Search and Rescue} Search and Rescue (SAR-$n$) was first introduced by Trevizan et al. \shortcite{trevizan16idual} as a \emph{Constrained} SSP. The goal is to find, board and escort to a safe location any one survivor in an $n\times n$ grid as quickly as possible, constrained by a fuel limit. Probabilities are introduced when modelling fuel consumption and partial observability of whether a survivor exists in a given location. The location of only one survivor is known for certain. We extend the problem by making fuel consumption as an additional objective instead of a constraint. A solution for a SAR MOSSP is a set of policies with different trade-offs between fuel and time.

\subsubsection{MO Triangle Tireworld} Triangle Tireworld, introduced by Little and Thi\'ebaux~\shortcite{little07probabilistic} as a probabilistically interesting problem, consists of a triangular grid of locations. The goal is to travel from an initial to a target location where each location has a probability of getting a flat tire. Some locations contain a spare tire which the agent can load into the car for future use.
These problems have dead ends and they occur when a tire is flat and no spare is available.
We apply the give-up transformation from Section~\ref{sec:movi} resulting in an MOSSP with two objectives: total number of actions and probability of using the give-up action.

\subsubsection{\ProbVisitall}
The deterministic MO VisitAll~\cite{geisser22moheuristics} is a TSP problem on a grid, where the agents must collectively visit all locations, and each agent's objective is to minimise its own number of moves.
This problem is considered MO interesting because any admissible ideal-point heuristic returns the zero vector for all states since it is possible for a single agent to visit all locations while no actions are performed by the other agents.
We extend this domain by adding a probabilistic action \emph{move-risky} which has probability 0.5 of acting as the regular move action and 0.5 of teleporting the agent back to its initial location.
The cost of the original move action was changed to 2 while the cost of the move-risky action is 1.

\subsubsection{\VisitallWithFlat} This domain combines features of both the probabilistic Tireworld and the deterministic MO VisitAll domains into one that is probabilistically and MO interesting.
The underlying structure and dynamics is the same as the deterministic MO VisitAll except that the move action now can result in a flat tire with probability 0.5.
We also added the actions for loading and changing tires for each of the agents.
Similarly to Triangle Tireworld, the problems in this domain can have dead ends when a flat tire occurs and no spare tire is available.
Applying the give-up transformation from Section~\ref{sec:movi} to this domain results in $k+1$ cost functions where $k$ is the number of agents.

\subsection{Setup}

We implemented the MO versions of the VI, TVI, (i)LAO$^*$ and LRTDP algorithms and the MO version of the PDB abstraction heuristics (\heur{mossp}{pdb}) in C++.\footnote{Code at \url{https://github.com/DillonZChen/cpp-mossp-planner}}
PDB heuristics are computed using TVI, $\e=0.001$ and $\vec{b}=\vec{100}$.
We include in our experiments the following heuristics for deterministic MO planning from~\cite{geisser22moheuristics}:
the ideal-point version of \hmax (\heur{ideal}{max});
the MO extension of \hmax (\heur{mo}{comax});
and the MO canonical PDBs of size 2 and 3 (\heur{mo}{pdb2} and \heur{mo}{pdb3}).
The SO PDB heuristics for SSPs from~\cite{klossner2021pdb} of size 2 and 3 (\heur{ideal}{pdb2} and \heur{ideal}{pdb3}) are also included in our experiments.
All problem configurations are run with a timeout of 1800 seconds, memory limit of 4GB, and a single CPU core.
The experiments are run on a cluster with Intel Xeon
3.2 GHz CPUs.
We used CPLEX version 22.1 as the LP solver for computing CCS.
The consistency threshold is set to $\e=0.001$ and we set $\vec{b}=\vec{100}$. Each experimental configuration is run 6 times and averages are taken to reduce variance in the data.

We also modify the inequality constraint in Alg.~4.3 from Roijers and Whiteson 2017 for computing CCS to
$\w \cdot (\vec{v} - \vec{v}') + x\leq -\lambda, \forall \vec{v}' \in \V$
with $\lambda = 0.01$ to deal with inevitable numerical precision errors when solving the LP.
If the $\lambda$ term is not added, the function may fail to prune unnecessary points such as those corresponding to stochastic policies in the CCS, resulting in slower convergence.
However, an overly large $\lambda$ may return incorrect results by discarding important points in the CCS.
One may alternatively consider the term as an MO parameter for trading off solution accuracy and search time, similarly to the $\e$ parameter.

\begin{figure}[t]
    \def\plotwidth{.92\linewidth}
        \centering 
        \includegraphics[width=\linewidth]{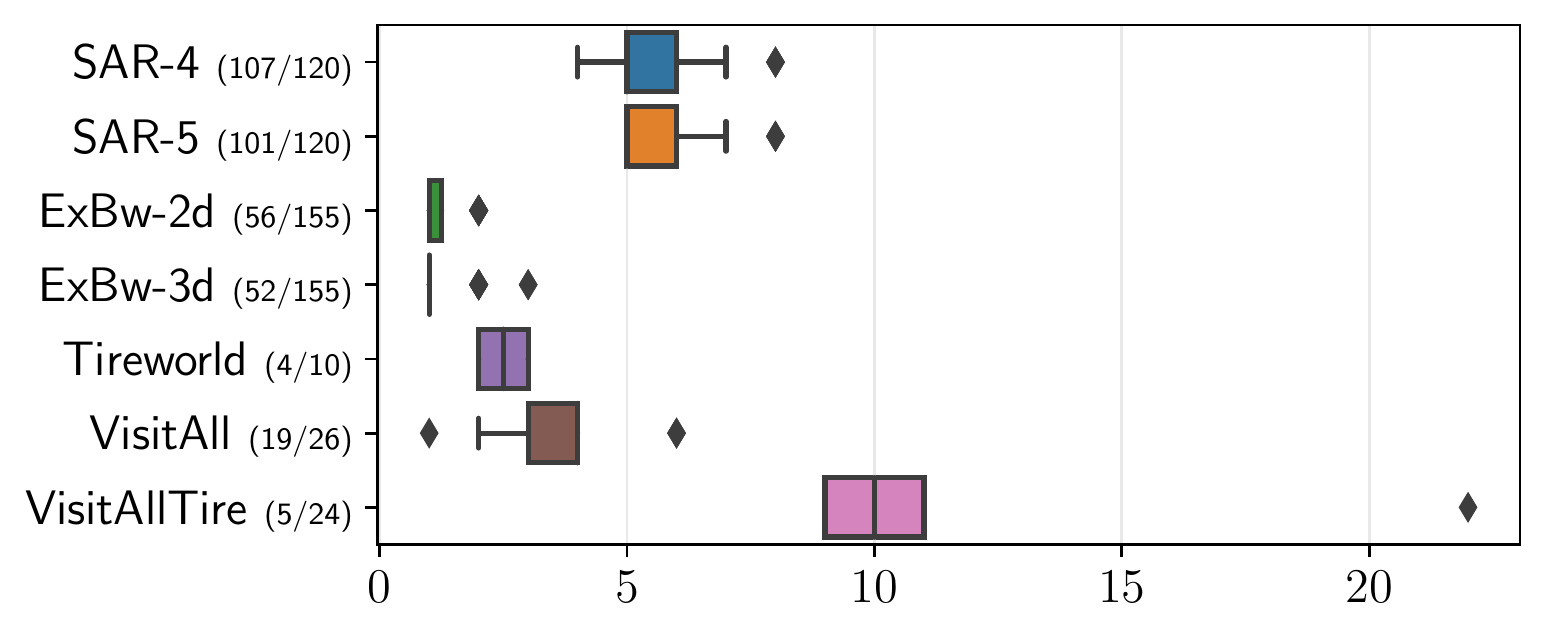}
    
    \caption{Boxplot of CCS size of instances across domains which have been solved at least once. Number of solved and total instances for each domain is indicated in parentheses.}
        \label{fig:ccs}
\end{figure}

Fig.~\ref{fig:ccs} shows the distribution of CCS sizes for different domains.
Recall that the CCS implicitly represents a potentially infinite set of stochastic policies and their value functions.
As a result, a small CCS is sufficient to dominate a large number of solutions, for instance, in Triangle Tireworld, a CCS of size 3 to 4 is enough to dominate all other policies even though the number of deterministic policies for this domain is double exponential in its parameter $n$.

\subsection{Results}

\begin{figure}[ht!]
\def\plotwidth{.97\linewidth}
\centering
    \begin{subfigure}[b]{\plotwidth}
    \includegraphics[width=\linewidth]{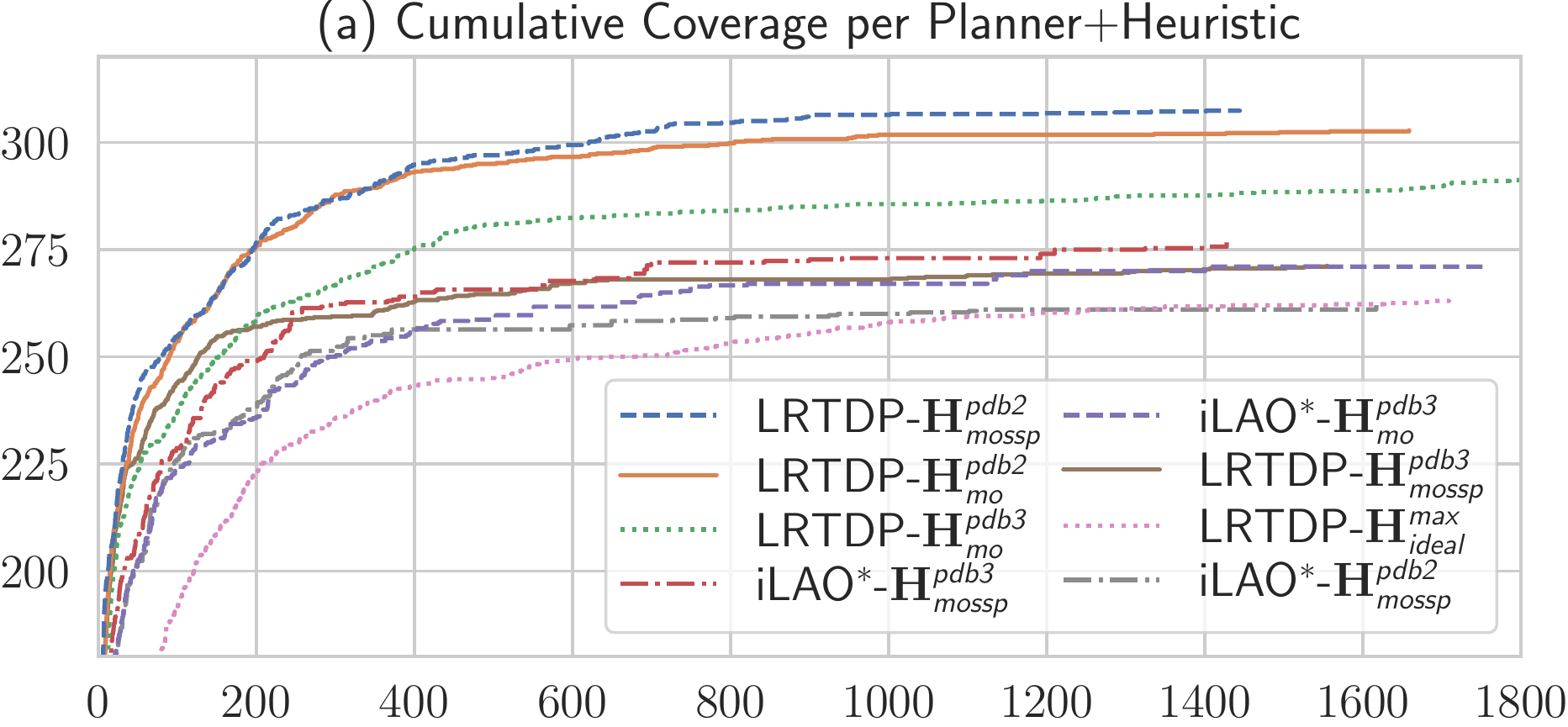}
    \setcounter{subfigure}{0}%
    \phantomcaption\label{fig:combo}
\end{subfigure}
\begin{subfigure}[b]{\plotwidth}
    \includegraphics[width=\linewidth]{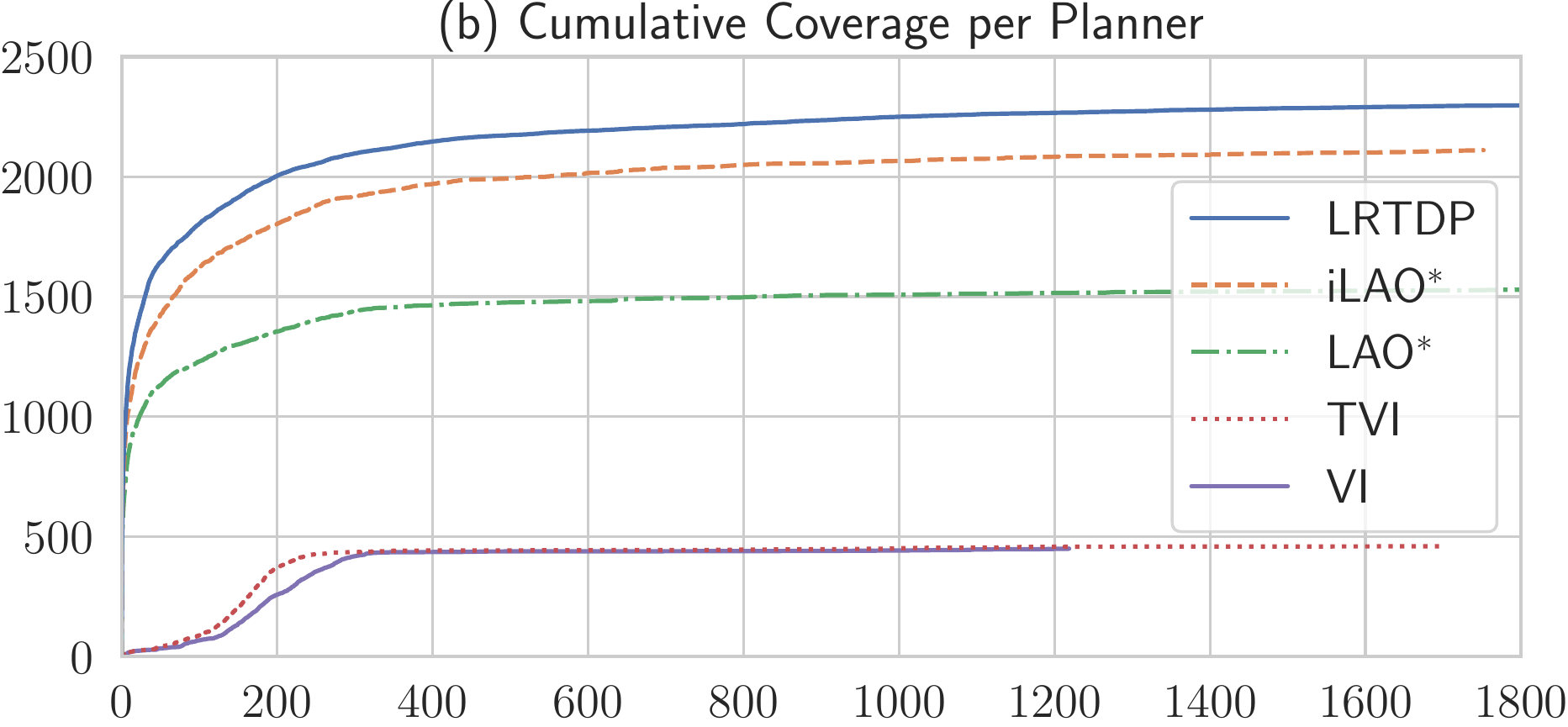}
    \phantomcaption\label{fig:planner}
\end{subfigure}
\begin{subfigure}[b]{\plotwidth}
    \includegraphics[width=\linewidth]{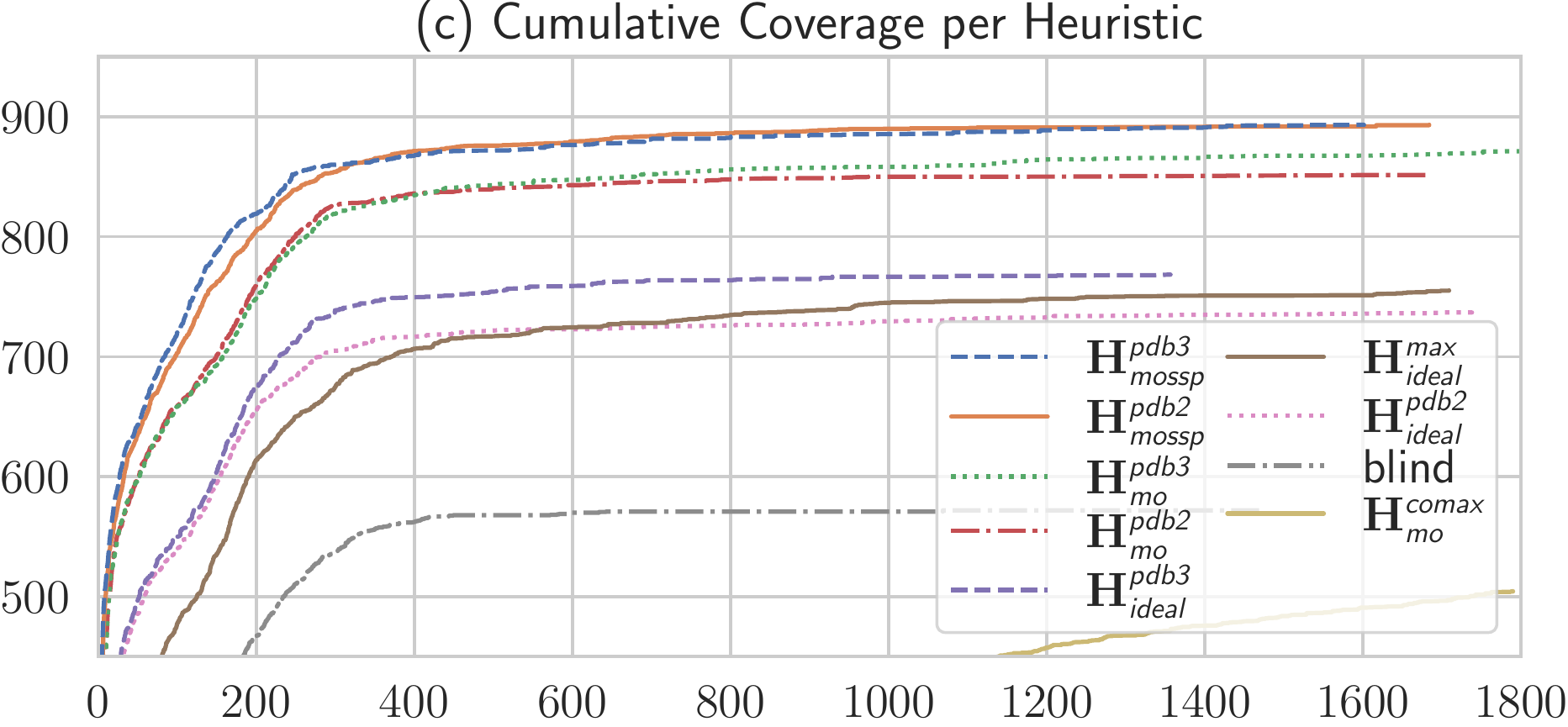}
    \phantomcaption\label{fig:heuristic}
\end{subfigure}
\begin{subfigure}[b]{\plotwidth}
    \includegraphics[width=\linewidth]{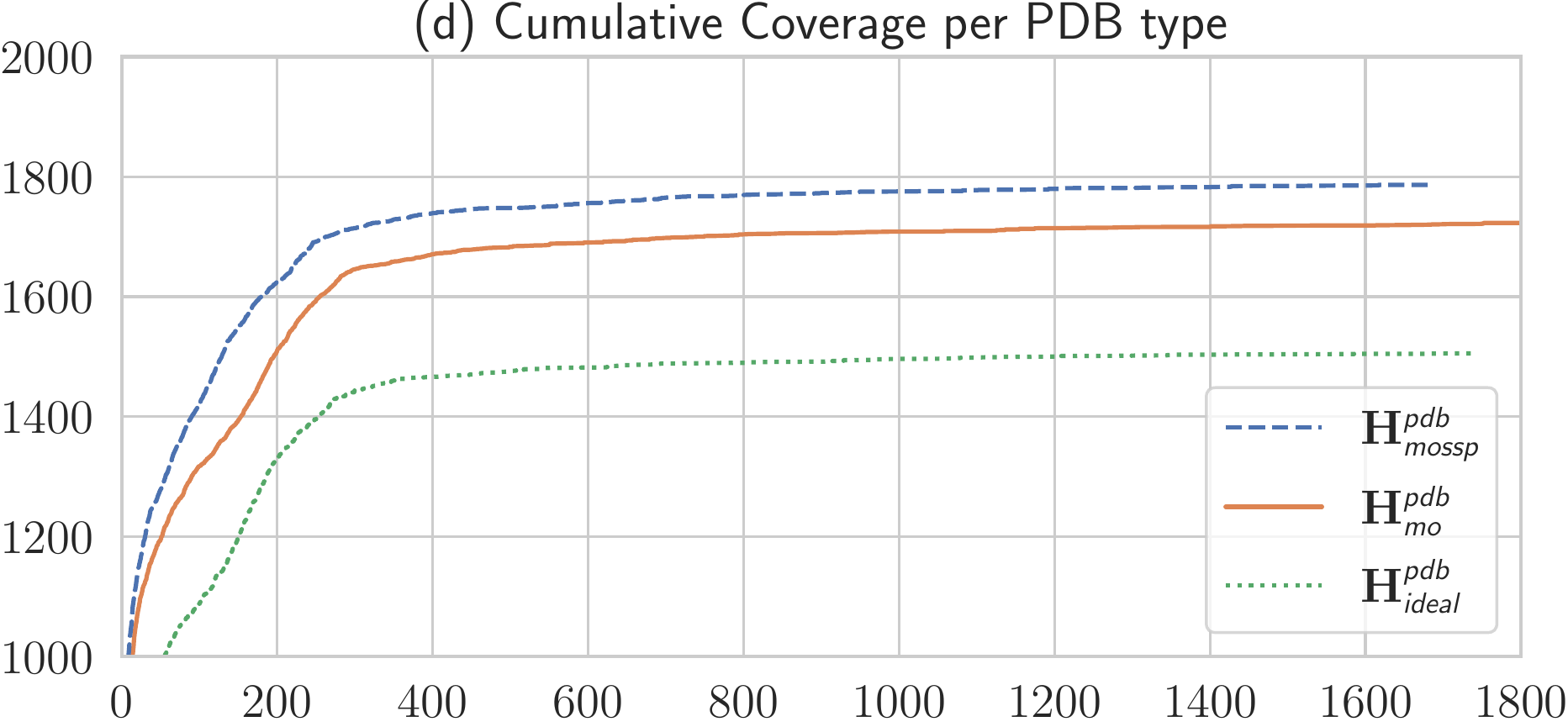}
    \phantomcaption\label{fig:pdb}
\end{subfigure}

\caption{Cumulative coverage of:
(a) planners and heuristics combinations (low-performing planners omitted);
(b) planners only, i.e., summation across different heuristics;
(c) heuristics only, i.e. summation across different planners;
(d) PDB approaches considered, i.e., summation across the ideal-point, MO, and MOSSP approaches.
Notice that the $x$-axis is the same for all plots but the $y$-axis is different and might not start
at 0.}
\label{fig:plots}
\end{figure}

\begin{table}[t]
  \centering
  {  
  \footnotesize
  {
  \begin{tabular}{L{0.9cm} L{0.9cm} C{1.6cm} L{0.9cm} C{1.6cm}}
  \cmidrule[\heavyrulewidth](lr){1-3}
  \cmidrule[\heavyrulewidth](lr){4-5}
  \multicolumn{2}{l}{planner + heuristic} & coverage & heuristic & coverage\\
  \cmidrule(lr){1-3}
  \cmidrule(lr){4-5}
  LRTDP&\heur{mossp}{pdb2} &$307.6\pm 2.7$    &\heur{mossp}{pdb3} &$893.5 \pm1.9$\\
  LRTDP&\heur{mo}{pdb2} &$302.8\pm 2.1$       &\heur{mossp}{pdb2} &$893.3 \pm3.2$\\
  LRTDP&\heur{mo}{pdb3} &$291.2\pm 1.3$       &\heur{mo}{pdb3} &$871.2 \pm1.8$\\
  iLAO$^*$&\heur{mossp}{pdb3} &$276.7\pm 0.7$ &\heur{mo}{pdb2} &$851.8 \pm2.1$\\
  iLAO$^*$&\heur{mo}{pdb3} &$272.0\pm 0.0$    &\heur{ideal}{pdb3} &$768.7 \pm0.7$\\
  LRTDP&\heur{mossp}{pdb3} &$271.2\pm 1.7$    &\heur{ideal}{max} &$755.2 \pm0.7$\\
  LRTDP&\heur{ideal}{max} &$263.2\pm 0.7$     &\heur{ideal}{pdb2} &$737.2 \pm1.2$\\
  iLAO$^*$&\heur{mo}{pdb2} &$262.0\pm 0.0$    &blind &$572.9 \pm1.4$\\
  \cmidrule[\heavyrulewidth](lr){1-3}
  \cmidrule[\heavyrulewidth](lr){4-5} \\[-0.4cm]
  \cmidrule[\heavyrulewidth](lr){1-3}
  \cmidrule[\heavyrulewidth](lr){4-5}
  planner & & coverage & PDB & coverage\\
  \cmidrule(lr){1-3}
  \cmidrule(lr){4-5}
  LRTDP & &$2297.3 \pm3.0$          & \heur{mossp}{pdb} &$1786.8\pm 1.6$\\
  iLAO$^*$ & &$2111.7 \pm0.7$       & \heur{mo}{pdb} &$1723.0\pm 2.0$\\
    LAO$^*$ & &$1529.0 \pm1.2$      & \heur{ideal}{pdb} &$1505.8\pm 1.9$\\
    \cmidrule[\heavyrulewidth](lr){4-5}
    \vspace*{-0.39cm}TVI & & \vspace*{-0.39cm}$460.0 \pm0.5$\\
    \vspace*{-0.35cm}VI  & & \vspace*{-0.35cm}$450.3 \pm0.7$\\[-0.1cm]
  \cmidrule[\heavyrulewidth](lr){1-3}
  \end{tabular}
  }}
  \caption{Average marginalised cumulative coverages with 95\% confidence intervals. Higher values are better.} \label{tab:coverage}
\end{table}

Fig.~\ref{fig:plots} summarises our results by showing the cumulative coverage of different
planners and heuristics across all considered problems. 
The following is a summary of our findings (we omit the MO in the planner names for simplicity):

\oldnb{ST: the name of the heuristics in the legend differs from that in the text.}

\subsubsection*{What is the best planner and heuristic combination?}
Referring to the overall cumulative coverage plot in Fig.~\ref{fig:combo} and coverage tables in Tab.~\ref{tab:coverage}, we notice that LRTDP+\heur{mossp}{pdb2} performs best followed by LRDTP+\heur{mo}{pdb2}, LRTDP+\heur{mo}{pdb3} and iLAO$^*$+\heur{mossp}{pdb3}.
The ranking of the top 3 planners remain the same if we normalise the coverage by domain.

We note that the considered Exbw3 problems had several improper policies resulting in slow convergence of TVI for solving PDBs of size 3.
This is due to Assumption~\ref{assump:weak} and the chosen $\vec{b}=\vec{100}$ for our experiments which that requires the cost of improper policies in each PDB to cost at least $100$ in one of its dimensions.
Ignoring the Exbw3 domain, the top 3 configurations are LRTDP+\heur{mossp}{pdb3}, LRTDP+\heur{mossp}{pdb2} and iLAO$^*$+\heur{mossp}{pdb3} and their 95\% confidence intervals all overlap.
The ranking of the top few planners remain the same if the coverage is normalised.

\subsubsection*{What planner is best?}

To answer this question, we look at the cumulative coverage marginalised over heuristics, that is, we sum the coverage of a planner across the different heuristics.
The results are shown in Fig.~\ref{fig:planner} and Tab.~\ref{tab:coverage} and the best three planners in order are LRTDP, iLAO$^*$ and LAO$^*$.
Notice that the difference in coverage between LRTDP and iLAO$^*$ is at $185.6$ solved instances while the difference between TVI and VI is only $9.7$.
The ranking between planners remains the same when the coverage is normalised per domain.
Considering domains individually, LRTDP is the best planner for Exbw and \VisitallWithFlat while for iLAO$^*$ is the best planner for SAR and \ProbVisitall.
For MO Tireworld, both LRTDP and iLAO$^*$ are tied in first place.

\newcommand{\hund}{{\color{gray}{100}}}
\newcommand{\othersize}{\footnotesize}
\newcommand{\benchmarkh}[2]{#1}
\newcommand{\tableheur}[2]{
  {\rotatebox[origin=l]{90}{{\!\!\othersize\raggedleft\heur{#1}{#2}}}}
}
\newcommand{\colorofcell}{blue}
\newcommand{\first}[1]{\cellcolor{\colorofcell!30}{\textbf{#1}}}
\newcommand{\second}[1]{\cellcolor{\colorofcell!20}{{#1}}}
\newcommand{\third}[1]{\cellcolor{\colorofcell!10}{{#1}}}
\setlength{\tabcolsep}{3.8pt}

\begin{table}[t]
\centering
{  
  \footnotesize
{
\begin{tabularx}{\linewidth}{l | r | r r | r r r r r r}
\toprule
\multicolumn{1}{c}{} & \multicolumn{1}{c}{} & \multicolumn{2}{c}{\othersize{Crit. Path}} & \multicolumn{6}{c}{\othersize{Abstractions}} \\
\multicolumn{1}{c}{} & \multicolumn{1}{c}{\rotatebox[origin=l]{90}{\!\!\othersize{blind}}} & \tableheur{ideal}{max} & \multicolumn{1}{c}{\tableheur{mo}{comax}}  & \tableheur{ideal}{pdb2} & \tableheur{ideal}{pdb3} & \tableheur{mo}{pdb2} & \tableheur{mo}{pdb3} & \tableheur{mossp}{pdb2} & \multicolumn{1}{c}{\tableheur{mossp}{pdb3}} \\
\toprule
\benchmarkh{SAR-4}{104} & \hund & 97 & \third{44} & 93 & 92 & \third{44} & \third{44} & \second{38} & \first{26}\\
\benchmarkh{SAR-5}{95} & \hund & 97 & \third{43} & 93 & 92 & \third{43} & \third{43} & \second{37} & \first{27}\\
\benchmarkh{ExBw-2d}{45} & \hund & 59 & \first{22} & 51 & \second{45} & 51 & \second{45} & 51 & \second{45}\\
\benchmarkh{ExBw-3d}{37} & \hund & 58 & \first{22} & 52 & \second{44} & 52 & \second{44} & 52 & \second{44}\\
\benchmarkh{Tireworld}{4} & \hund & \hund & \third{68} & \hund & \hund & \third{68} & \third{68} & \second{57} & \first{13}\\
\benchmarkh{VisitAll}{19} & \hund & \hund & 54 & \hund & \hund & 61 & \third{53} & \second{22} & \first{12}\\
\benchmarkh{VisitAllTire}{5} & \hund & \hund & \third{50} & \hund & \hund & 66 & \first{45} & 66 & \first{45}\\
\bottomrule
\end{tabularx}
}}
\caption{The mean relative error (\%) of the heuristic value relative to the optimal value at the initial state computed with the \emph{directed} Hausdorff distance divided by the norm of the largest vector in the optimal value: $\max_{\vec{v} \in \V^*}\min_{\vec{u} \in \H} d(\vec{v}, \vec{u}) / \max_{\vec{v} \in \V^*} \|\vec{v}\|$. Only solved instances for which all heuristics were computed for the initial state within the time limit were considered. Cells ranked first to third are shaded. Lower values are better.} \label{tab:h}
\end{table}

\subsubsection*{What heuristic is best?}
The cumulative coverage marginalised over planners for selected heuristics is shown in Fig.~\ref{fig:heuristic} and Tab.~\ref{tab:coverage}. The MOSSP PDB heuristic of size 3 (\heur{mossp}{pdb3}) has the best performance followed by \heur{mossp}{pdb2}, \heur{mo}{pdb3} and \heur{mo}{pdb2}.
We note there is a large overlap in the 95\% confidence intervals of \heur{mossp}{pdb2} and \heur{mossp}{pdb3} due to the slow computation of \heur{mossp}{pdb3} on Exbw3. The overlap disappears when we remove the Exbw3 results with coverage and confidence intervals given by $862.9 \pm3.0$ and $807.6 \pm2.9$ for \heur{mossp}{pdb3} and \heur{mossp}{pdb2}, respectively.
We further quantify heuristic accuracy using the directed Hausdorff distance between heuristic and optimal values in Tab.~\ref{tab:h}. We notice that \heur{mo}{comax} achieves strong accuracy relative to other heuristics but has the worst coverage of $504.5$ due to its high computation time.

\subsubsection*{What feature of MOSSPs is more important to capture in a heuristic?}
To answer this question, consider the performance of the 3 classes of PDB heuristics: probabilistic-only PDBs (\heur{ideal}{pdb}), MO only PDBs (\heur{mo}{pdb}), and MO probabilistic PDBs (\heur{mossp}{pdb}).
The cumulative coverage marginalised over the different PDB heuristics shown in Fig.~\ref{fig:pdb} and Tab.~\ref{tab:coverage} highlights the effectiveness of MO PDBs.
\heur{mossp}{pdb} is able to solve $63.8$ and $281$ more instances than \heur{mo}{pdb} and \heur{ideal}{pdb}, respectively.
The ranking remains the same when the coverage is normalised per domain.
Moreover, notice in Tab.~\ref{tab:h} that \heur{mo}{pdb} heuristics are at least as informative as \heur{ideal}{pdb} heuristics on all domains.
Lastly, we note that an  admissible ideal point heuristic is upper bounded by $\{\uu\}$ where $\uu_i = \min_{\vv \in \V^*} \vv_i$ for $i=1,\ldots,n$.
These results suggest that it is more important for a heuristic to maintain the MO cost structure of MOSSPs than the stochastic structure of actions.

\section{Conclusion}

In this work we define a new general class of problems, namely multi-objective stochastic shortest path problems (MOSSPs).
We adapt MOVI which was originally constructed for solving multi-objective Markov Decision Processes to our MOSSP setting with conditions for convergence.
We further design new, more powerful heuristic search algorithms (i)MOLAO$^*$ and MOLRTDP for solving MOSSPs.
The algorithms are complemented by a range of MOSSP heuristics and an extensive set of experiments on several benchmarks with varying difficulty and features.
Through our evaluation, we can conclude that MOLRTDP is the best performing MOSSP solver, and abstraction heuristics which consider both the MO and probabilistic aspects of MOSSPs the best performing MOSSP heuristic.
Our future work includes adding probabilistic LTL constraints as additional multi-objectives in order to compute solutions to MO-PLTL SSPs~\cite{baumgartner18:pltldual} that are robust to the choice of probability thresholds.

\section*{Acknowledgements}

This work was supported by ARC project DP180103446, \emph{On-line planning for constrained autonomous agents in an uncertain world}.
The computational resources for this project were provided by the Australian Government through the National Computational Infrastructure (NCI) under the ANU Startup Scheme.

\bibliography{bib}

\begin{thebibliography}{27}
\providecommand{\natexlab}[1]{#1}

\bibitem[{Barrett and Narayanan(2008)}]{barrett08learning}
Barrett, L.; and Narayanan, S. 2008.
\newblock Learning All Optimal Policies with Multiple Criteria.
\newblock In \emph{Proc. 25th International Conference on Machine Learning
  (ICML)}, 41--47.

\bibitem[{Baumgartner, Thi{\'e}baux, and
  Trevizan(2018)}]{baumgartner18:pltldual}
Baumgartner, P.; Thi{\'e}baux, S.; and Trevizan, F. 2018.
\newblock {H}euristic {S}earch {P}lanning {W}ith {M}ulti-{O}bjective
  {P}robabilistic {LTL} {C}onstraints.
\newblock In \emph{Proc. of 16th Int. Conf. on Principles of Knowledge
  Representation and Reasoning (KR)}.

\bibitem[{Bertsekas and Tsitsiklis(1991)}]{bertsekas1991analysis}
Bertsekas, D.~P.; and Tsitsiklis, J.~N. 1991.
\newblock An Analysis of Stochastic Shortest Path Problems.
\newblock \emph{Mathematics of Operations Research}, 16(3): 580--595.

\bibitem[{Bertsekas and Tsitsiklis(1996)}]{bertsekas96neuro}
Bertsekas, D.~P.; and Tsitsiklis, J.~N. 1996.
\newblock \emph{Neuro-Dynamic Programming}, volume~3 of \emph{Optimization and
  neural computation series}.
\newblock Athena scientific.

\bibitem[{Bonet and Geffner(2001)}]{bonet2001heuristics}
Bonet, B.; and Geffner, H. 2001.
\newblock Planning as Heuristic Search.
\newblock \emph{Artif. Intell.}, 129(1-2): 5--33.

\bibitem[{Bonet and Geffner(2003)}]{bonet2003lrtdp}
Bonet, B.; and Geffner, H. 2003.
\newblock Labeled {RTDP:} Improving the Convergence of Real-Time Dynamic
  Programming.
\newblock In \emph{Proc. 13th International Conference on Automated Planning
  and Scheduling (ICAPS)}, 12--21.

\bibitem[{Bryce and Buffet(2008)}]{bryce08ippc}
Bryce, D.; and Buffet, O. 2008.
\newblock 6th {I}nt. {P}lanning {C}ompetition: {U}ncertainty {T}rack.
\newblock \emph{3rd Int. Probabilistic Planning Competition}.

\bibitem[{Bryce, Cushing, and Kambhampati(2007)}]{bryce2007probabilistic}
Bryce, D.; Cushing, W.; and Kambhampati, S. 2007.
\newblock Probabilistic Planning is Multi-Objective.
\newblock Technical Report 07-006, ASU CSE.

\bibitem[{Dai et~al.(2014)Dai, Mausam, Weld, and
  Goldsmith}]{dai2014topological}
Dai, P.; Mausam; Weld, D.~S.; and Goldsmith, J. 2014.
\newblock Topological Value Iteration Algorithms.
\newblock \emph{J. Artif. Intell. Res.}, 42: 181--209.

\bibitem[{Gei{\ss}er et~al.(2022)Gei{\ss}er, Haslum, Thi{\'e}baux, and
  Trevizan}]{geisser22moheuristics}
Gei{\ss}er, F.; Haslum, P.; Thi{\'e}baux, S.; and Trevizan, F. 2022.
\newblock Admissible Heuristics for Multi-Objective Planning.
\newblock In \emph{Proc. 32nd International Conference on Automated Planning
  and Scheduling (ICAPS)}, 100--109.

\bibitem[{Hansen and Zilberstein(2001)}]{hansen01lao}
Hansen, E.~A.; and Zilberstein, S. 2001.
\newblock {LAO}\({}^{\mbox{*}}\): {A} Heuristic search algorithm that finds
  solutions with loops.
\newblock \emph{Artif. Intell.}, 129(1-2): 35--62.

\bibitem[{Jimenez, Coles, and Smith(2006)}]{jimenez06determinisation}
Jimenez, S.; Coles, A.; and Smith, A. 2006.
\newblock Planning in Probabilistic Domains using a Deterministic Numeric
  Planner.
\newblock In \emph{Proc. 25th Workshop of the UK Planning and Scheduling
  Special Interest Group (PLANSIG)}, 74--79.

\bibitem[{Keller and Helmert(2013)}]{keller2013trial}
Keller, T.; and Helmert, M. 2013.
\newblock Trial-Based Heuristic Tree Search for Finite Horizon MDPs.
\newblock In \emph{Proc. 23rd International Conference on Automated Planning
  and Scheduling ({ICAPS})}. {AAAI}.

\bibitem[{Khouadjia et~al.(2013)Khouadjia, Schoenauer, Vidal, Dr\'eo, and
  Sav\'eant}]{Khouadjia2013}
Khouadjia, M.; Schoenauer, M.; Vidal, V.; Dr\'eo, J.; and Sav\'eant, P. 2013.
\newblock Pareto-Based Multiobjective {AI} Planning.
\newblock In \emph{Proc. 23rd International Joint Conference on Artificial
  Intelligence (IJCAI)}, 2321--2327.

\bibitem[{Kl{\"{o}}{\ss}ner and Hoffmann(2021)}]{klossner2021pdb}
Kl{\"{o}}{\ss}ner, T.; and Hoffmann, J. 2021.
\newblock Pattern Databases for Stochastic Shortest Path Problems.
\newblock In Ma, H.; and Serina, I., eds., \emph{Proc.\ 14th International
  Symposium on Combinatorial Search (SOCS)}, 131--135.

\bibitem[{Little and Thi{\'e}baux(2007)}]{little07probabilistic}
Little, I.; and Thi{\'e}baux, S. 2007.
\newblock {P}robabilistic {P}lanning vs {R}eplanning.
\newblock In \emph{Proc.\ {ICAPS} {W}orkshop {I}nternational {P}lanning
  {C}ompetition: {P}ast, {P}resent and {F}uture}.

\bibitem[{Mandow and P{\'{e}}rez{-}de{-}la{-}Cruz(2010)}]{mandow2010namoa}
Mandow, L.; and P{\'{e}}rez{-}de{-}la{-}Cruz, J. 2010.
\newblock Multiobjective A\(^*\) search with consistent heuristics.
\newblock \emph{J. {ACM}}, 57(5): 27:1--27:25.

\bibitem[{Painter, Lacerda, and Hawes(2020)}]{painter2020convex}
Painter, M.; Lacerda, B.; and Hawes, N. 2020.
\newblock Convex Hull Monte-Carlo Tree-Search.
\newblock In \emph{Proc. 30th International Conference on Automated Planning
  and Scheduling ({ICAPS})}, 217--225.

\bibitem[{Roijers and Whiteson(2017)}]{roijers17modm}
Roijers, D.~M.; and Whiteson, S. 2017.
\newblock \emph{Multi-Objective Decision Making}.
\newblock Synthesis Lectures on Artificial Intelligence and Machine Learning.
  Morgan {\&} Claypool Publishers.

\bibitem[{Trevizan, Teichteil-K{\"o}nigsbuch, and
  Thi{\'e}baux(2017)}]{trevizan17:mcmp}
Trevizan, F.; Teichteil-K{\"o}nigsbuch, F.; and Thi{\'e}baux, S. 2017.
\newblock {E}fficient {S}olutions for {S}tochastic {S}hortest {P}ath {P}roblems
  with {D}ead {E}nds.
\newblock In \emph{Proc. 33rd International Conference on Uncertainty in
  Artificial Intelligence ({UAI})}.

\bibitem[{Trevizan et~al.(2017)Trevizan, Thi{\'e}baux, Santana, and
  Williams}]{trevizan16idual}
Trevizan, F.; Thi{\'e}baux, S.; Santana, P.; and Williams, B. 2017.
\newblock {I}-dual: {S}olving {C}onstrained {SSP}s via {H}euristic {S}earch in
  the {D}ual {S}pace.
\newblock In \emph{Proc.\ 26th International Joint Conference on Artificial
  Intelligence (IJCAI)}.

\bibitem[{Trevizan, Thi{\'{e}}baux, and Haslum(2017)}]{trevizan17hpom}
Trevizan, F.~W.; Thi{\'{e}}baux, S.; and Haslum, P. 2017.
\newblock Occupation Measure Heuristics for Probabilistic Planning.
\newblock In \emph{Proc.\ 27th International Conference on Automated Planning
  and Scheduling (ICAPS)}, 306--315.

\bibitem[{Ulloa et~al.(2020)Ulloa, Yeoh, Baier, Zhang, Suazo, and
  Koenig}]{ulloa2020boastar}
Ulloa, C.~H.; Yeoh, W.; Baier, J.~A.; Zhang, H.; Suazo, L.; and Koenig, S.
  2020.
\newblock A Simple and Fast Bi-Objective Search Algorithm.
\newblock In \emph{Proc.\ 30th International Conference on Automated Planning
  and Scheduling (ICAPS)}, 143--151.

\bibitem[{White(1982)}]{white82mdp}
White, D.~J. 1982.
\newblock Multi-objective infinite-horizon discounted Markov decision
  processes.
\newblock \emph{Journal of Mathematical Analysis and Applications}, 89(2):
  639--647.

\bibitem[{Wiering and de~Jong(2007)}]{wiering07conmodp}
Wiering, M.~A.; and de~Jong, E.~D. 2007.
\newblock Computing Optimal Stationary Policies for Multi-Objective Markov
  Decision Processes.
\newblock In \emph{Proc.\ 1st IEEE International Symposium on Approximate
  Dynamic Programming and Reinforcement Learning}, 158--165.

\bibitem[{Wingate and Seppi(2005)}]{wingate2005prioritization}
Wingate, D.; and Seppi, K.~D. 2005.
\newblock Prioritization Methods for Accelerating {MDP} Solvers.
\newblock \emph{J. Mach. Learn. Res.}, 6: 851--881.

\bibitem[{Younes et~al.(2005)Younes, Littman, Weissman, and
  Asmuth}]{younes2005competition}
Younes, H. L.~S.; Littman, M.~L.; Weissman, D.; and Asmuth, J. 2005.
\newblock The First Probabilistic Track of the International Planning
  Competition.
\newblock \emph{J. Artif. Intell. Res.}, 24: 851--887.

\end{thebibliography}
\clearpage
\appendix
\renewcommand{\othersize}{\small}
\newcommand{\benchmark}[2]{
  {\multirow{5}{*}{\rotatebox[origin=c]{90}{#1}}} & 
  {\multirow{5}{*}{\rotatebox[origin=c]{90}{(#2)}}}
}
\newcommand{\solver}[1]{{\textsc{#1}}}
\renewcommand{\tableheur}[2]{{\othersize\heur{#1}{#2}}}
\begin{table*}[!ht]  
\centering
\textbf{\Huge Appendix} \vspace*{1cm}\\
\begin{tabular}{l l l | r | r r | r r r r r r | r c}
\toprule
& & \multicolumn{1}{c}{} & \multicolumn{1}{c}{} & \multicolumn{2}{c}{{Critical Path}} & \multicolumn{6}{c}{{Abstractions}} \\
& & \multicolumn{1}{c}{} & \multicolumn{1}{c}{\othersize{blind}} & \tableheur{ideal}{max} & \multicolumn{1}{c}{\tableheur{mo}{comax}} & \tableheur{ideal}{pdb2} & \tableheur{ideal}{pdb3} & \tableheur{mo}{pdb2} & \tableheur{mo}{pdb3} & \tableheur{mossp}{pdb2} & \multicolumn{1}{c}{\tableheur{mossp}{pdb3}} & \multicolumn{1}{c}{\othersize{Sum}} \\
\toprule
\benchmark{SAR-4}{120}
& \solver{VI} & 39.7 & 39.0 & - & 39.0 & 39.0 & 35.0 & 38.0 & 39.0 & 39.0 & 307.7 \\
& & \solver{TVI} & 40.0 & 40.0 & - & 40.0 & 40.0 & 40.0 & 40.0 & 40.0 & 40.0 & 320.0 \\
& & \solver{LAO$^*$} & 51.0 & 54.0 & 83.0 & 62.0 & 66.0 & 76.0 & 77.0 & 84.3 & 91.0 & 644.3 \\
& & \solver{iLAO$^*$} & 80.0 & 90.0 & 75.0 & 94.0 & 100.0 & 90.0 & 97.0 & 93.0 & 104.0 & 823.0 \\
& & \solver{LRTDP} & 72.2 & 90.8 & 80.8 & 93.8 & 95.8 & 91.0 & 93.6 & 98.2 & 99.4 & 815.5 \\ \midrule
\benchmark{SAR-5}{120}
& \solver{VI} & - & - & - & - & - & - & - & - & 0.7 & 0.7 \\
& & \solver{TVI} & - & - & - & - & - & - & - & - & 0.7 & 0.7 \\
& & \solver{LAO$^*$} & 42.0 & 38.0 & 54.0 & 54.0 & 58.0 & 71.0 & 76.7 & 82.0 & 87.7 & 563.3 \\
& & \solver{iLAO$^*$} & 65.0 & 81.0 & 43.0 & 84.0 & 97.0 & 88.0 & 96.0 & 94.0 & 100.0 & 748.0 \\
& & \solver{LRTDP} & 55.5 & 80.8 & 41.2 & 86.0 & 89.5 & 88.0 & 91.0 & 91.2 & 93.8 & 717.0 \\ \midrule
\benchmark{ExBw-2d}{155}
& \solver{VI} & - & - & - & - & - & - & - & - & - & - \\
& & \solver{TVI} & - & - & - & - & - & - & - & - & - & - \\
& & \solver{LAO$^*$} & 4.0 & 29.0 & 10.0 & 7.7 & 11.7 & 18.0 & 20.3 & 16.3 & 22.3 & 139.3 \\
& & \solver{iLAO$^*$} & 5.0 & 26.0 & 8.0 & 17.0 & 21.0 & 31.0 & 32.0 & 30.0 & 32.0 & 202.0 \\
& & \solver{LRTDP} & 22.0 & 39.8 & 11.5 & 33.0 & 27.0 & 52.8 & 44.2 & 47.6 & 42.2 & 320.1 \\ \midrule
\benchmark{ExBw-3d}{155}
& \solver{VI} & - & - & - & - & - & - & - & - & - & - \\
& & \solver{TVI} & - & - & - & - & - & - & - & - & - & - \\
& & \solver{LAO$^*$} & 2.0 & 21.0 & 9.0 & 10.0 & 9.0 & 15.0 & 17.3 & 14.7 & 7.3 & 105.3 \\
& & \solver{iLAO$^*$} & 5.0 & 21.0 & 8.0 & 16.0 & 20.0 & 28.0 & 27.0 & 26.0 & 14.7 & 165.7 \\
& & \solver{LRTDP} & 20.0 & 36.0 & 11.0 & 31.5 & 26.0 & 50.0 & 41.4 & 45.0 & 8.6 & 269.5 \\ \midrule
\benchmark{Tireworld}{10}
& \solver{VI} & 3.0 & 3.0 & 2.0 & 3.0 & 3.0 & 3.0 & 3.0 & 3.0 & 3.0 & 26.0 \\
& & \solver{TVI} & 2.0 & 2.0 & 2.0 & 3.0 & 3.0 & 3.0 & 2.7 & 3.0 & 3.0 & 23.7 \\
& & \solver{LAO$^*$} & 2.0 & 2.0 & 2.0 & 2.0 & 2.0 & 2.0 & 2.0 & 2.0 & 3.0 & 19.0 \\
& & \solver{iLAO$^*$} & 3.0 & 3.0 & 2.0 & 3.0 & 3.0 & 3.0 & 3.0 & 3.0 & 4.0 & 27.0 \\
& & \solver{LRTDP} & 3.0 & 3.0 & 2.0 & 3.0 & 3.0 & 3.0 & 3.0 & 3.0 & 4.0 & 27.0 \\ \midrule
\benchmark{Pr. VisitAll}{26}
& \solver{VI} & 13.0 & 13.0 & 12.0 & 13.0 & 13.0 & 12.0 & 12.0 & 13.0 & 15.0 & 116.0 \\
& & \solver{TVI} & 13.0 & 13.0 & 12.0 & 13.0 & 13.0 & 12.0 & 12.0 & 13.7 & 14.0 & 115.7 \\
& & \solver{LAO$^*$} & 2.0 & 2.0 & 6.0 & 1.0 & 1.0 & 5.0 & 6.0 & 11.7 & 18.0 & 52.7 \\
& & \solver{iLAO$^*$} & 14.0 & 14.0 & 14.0 & 14.0 & 14.0 & 15.0 & 14.0 & 15.0 & 19.0 & 133.0 \\
& & \solver{LRTDP} & 10.5 & 10.2 & 13.0 & 10.2 & 9.8 & 13.0 & 13.0 & 17.6 & 18.2 & 115.5 \\ \midrule
\benchmark{VisitAllTire}{24}
& \solver{VI} & - & - & - & - & - & - & - & - & - & - \\
& & \solver{TVI} & - & - & - & - & - & - & - & - & - & - \\
& & \solver{LAO$^*$} & - & - & 1.0 & - & - & 1.0 & 1.0 & 1.0 & 1.0 & 5.0 \\
& & \solver{iLAO$^*$} & 1.0 & 1.0 & 1.0 & 1.0 & 1.0 & 1.0 & 3.0 & 1.0 & 3.0 & 13.0 \\
& & \solver{LRTDP} & 3.0 & 2.8 & 1.0 & 3.0 & 3.0 & 5.0 & 5.0 & 5.0 & 5.0 & 32.8 \\ \midrule
\benchmark{Sum}{610}
& \solver{VI} & 55.7 & 55.0 & 14.0 & 55.0 & 55.0 & 50.0 & 53.0 & 55.0 & 57.7 & 450.3 \\
& & \solver{TVI} & 55.0 & 55.0 & 14.0 & 56.0 & 56.0 & 55.0 & 54.7 & 56.7 & 57.7 & 460.0 \\
& & \solver{LAO$^*$} & 103.0 & 146.0 & 165.0 & 136.7 & 147.7 & 188.0 & 200.3 & 212.0 & 230.3 & 1529.0 \\
& & \solver{iLAO$^*$} & 173.0 & 236.0 & 151.0 & 229.0 & 256.0 & 256.0 & 272.0 & 262.0 & 276.7 & 2111.7 \\
& & \solver{LRTDP} & 186.2 & 263.2 & 160.5 & 260.5 & 254.0 & 302.8 & 291.2 & 307.6 & 271.2 & 2297.3 \\ \bottomrule
\end{tabular}
\caption{Number of tasks solved with different MOSSP solvers and heuristics.  SAR-5 increases in difficulty from SAR-4 with the increase in search space size. ExBw-3d increases in difficulty from ExBw-2d in the number of objectives. The rightmost sum column marginalises over heuristics, whereas the bottom sum rows marginalises over domains.} \label{tab:num_solved}
\end{table*} 

\end{document}